\documentclass[letterpaper, 10 pt, conference]{ieeeconf}  

\IEEEoverridecommandlockouts                              

\usepackage{cite}
\usepackage{amsmath,amssymb,amsfonts}
\usepackage{algorithmic}
\usepackage{graphicx}
\usepackage{textcomp}
\usepackage{xcolor}
\usepackage{hyperref}
\usepackage{multirow}
\usepackage{balance}

\title{\LARGE \bf
RATLIP: Generative Adversarial CLIP Text-to-Image Synthesis Based on Recurrent Affine Transformations
}

\author{Chengde Lin, Xijun Lu and Guangxi Chen
\thanks{This work was in part supported by National college students' innovation and entrepreneurship training program(Grant No.202310595062) and Guangxi college students' innovation and entrepreneurship training program(Grant No.S202310595240).(Corresponding author: Guangxi Chen.)}%
\thanks{Chengde Lin, Xijun Lu and Guangxi Chen are with 
School of Artificial Intelligence, Guangxi Colleges and Universities Key Laboratory of AI Algorithm Engineering, Guilin University of Electronic Technology, Guilin 541004, Guangxi, China(e-mail:lcd@guet.edu.cn, 2101620211@mails.guet.edu.cn, chgx@guet.edu.cn).}%
\thanks{Chengde Lin and Xijun Lu contribute equally to this paper.}
}
\begin{document}

\maketitle
\thispagestyle{empty}
\pagestyle{empty}

\begin{abstract}
Synthesizing high-quality photorealistic images with textual descriptions as a condition is very challenging. Generative Adversarial Networks (GANs), the classical model for this task, frequently suffer from low consistency between image and text descriptions and insufficient richness in synthesized images. Recently, conditional affine transformations (CAT), such as conditional batch normalization and instance normalization, have been applied to different layers of GAN to control content synthesis in images. CAT is a multi-layer perceptron that independently predicts data based on batch statistics between neighboring layers, with global textual information unavailable to other layers. To address this issue, we first model CAT and a recurrent neural network (RAT) to ensure that different layers can access global information. We then introduce shuffle attention between RAT to mitigate the characteristic of information forgetting in recurrent neural networks. Moreover, both our generator and discriminator utilize the powerful pre-trained model, Clip, which has been extensively employed for establishing associations between text and images through the learning of multimodal representations in latent space. The discriminator utilizes CLIP's ability to comprehend complex scenes to accurately assess the quality of the generated images. Extensive experiments have been conducted on the CUB, Oxford, and CelebA-tiny datasets to demonstrate the superiority of the proposed model over current state-of-the-art models. The code is \href{https://github.com/OxygenLu/RATLIP}{https://github.com/OxygenLu/RATLIP}.

\end{abstract}

\section{INTRODUCTION}
Cross-modal generative artificial intelligence is growing rapidly. Among these, text-to-image synthesis has emerged as a bridge between computer vision and natural language processing. Using descriptive sentences as conditions for synthesizing images is a highly prominent research topic. It has numerous potential application scenarios, including text-driven image editing, virtual picture synthesis, and face reconstruction, among others. Many methods have been proposed to fulfill the application requirements. Recently, based on a large number of datasets, model parametric quantities and pre-training weights, the diffusion models \cite{Xie_2023_ICCV} \cite{2024diffusion}\cite{xue2024} have achieved impressive results in text-to-image synthesis. They are capable of synthesizing images of complex scenes and outperforming Generative Adversarial Network (GAN)-based models in the same task, yielding satisfactory picture synthesis. However, they are consistently plagued by the problem of long inference times and high arithmetic overhead.

Recently, GAN-based approaches have given hope to solve the above problem. Tao et al. \cite{tao2023galip} proposed Generative Adversarial CLIPs for Text-to-Image Synthesis (GALIP), wherein they utilize frozen CLIP weights to guide text-to-synthesized image generation and employ discriminators for adversarial loss. During inference, the model can synthesize a high-quality image every 0.04 seconds. Additionally, the model is designed to alleviate the instability of GANs during training, which are prone to pattern collapse.

Text-to-image synthesis using GANs, encoded text information is adaptively fused into the model through dedicated fusion blocks. Common methods of text information fusion include Conditional Instance Normalization (CIN) and Conditional Bulk Normalization (CBN).
Initially, Conditional Instance Normalization (CIN) was proposed for style migration \cite{dumoulin2016learned}. With the help of CIN, StyleGAN\cite{karras2019style} synthesized impressive high-quality images with style migration. In recent years, with the development of cross-modal text-to-image synthesis, it has been discovered that CIN can also be applied to text information fusion, as seen in approaches like Styleclip\cite{patashnik2021styleclip}, to achieve text-driven image synthesis.
BigGAN\cite{brock2018large} is a typical model based on Conditional Bulk Normalization (CBN) synthesis. Recently, DF-GAN\cite{tao2022df} and GALIP\cite{tao2023galip} have been inspired by Conditional Bulk Normalization (CBN) and AdaIN\cite{huang2017arbitrary} to design the DFBlock aiming to fully exploit text information during text fusion.

\begin{figure*}[!ht]
\centering
\includegraphics[width=6.5in]{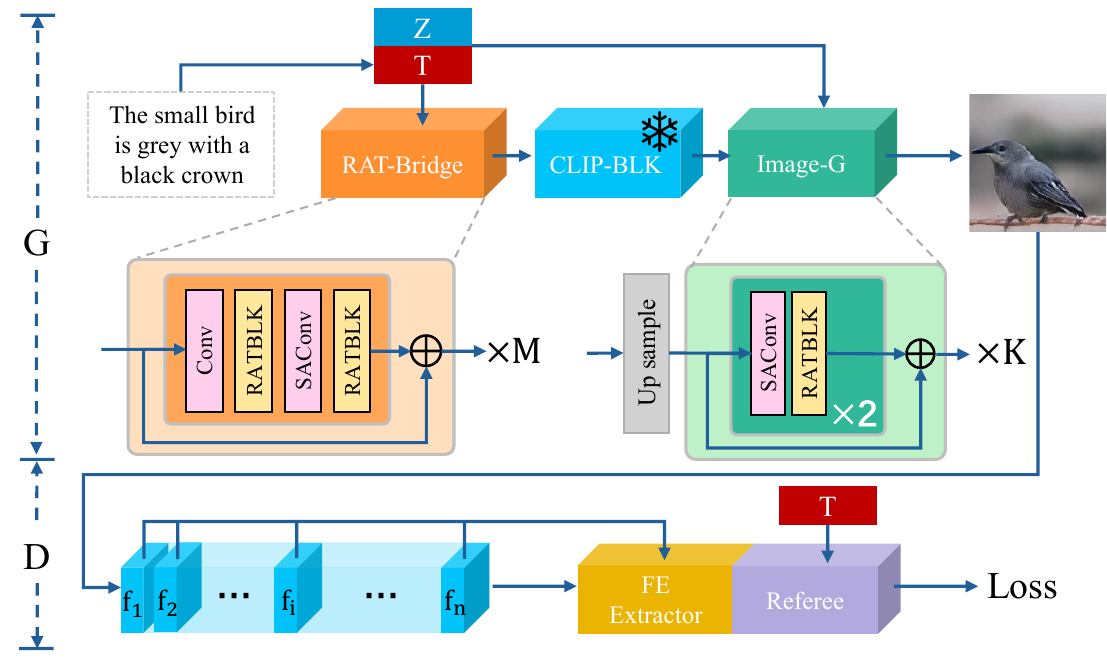}
\caption{The proposed general framework, RATLIP, for text-to-image synthesis. By integrating RATBLK and SAConv into the generator, our model can effectively understand the context.}
\label{all}
\end{figure*}
Despite the popularity of these text fusion approaches, they suffer from a serious drawback. Isolated feature fusion blocks allow Conditional Instance Normalizations (CINs) to occur independently in different layers, disregarding the semantic relationships of the fused text information across layers and within the global text information. Moreover, isolated fusion blocks are challenging to optimize as they are perceived to not interact with each other in the model.

This paper proposes an adversarial CLIP text-to-image synthesis approach based on recurrent affine transformations. We utilize a specialized recurrent neural network (LSTM) to model the RAT-Block, ensuring consistent batch statistics between fused blocks by employing jump connections and weight sharing of LSTM. The fusion block is no longer an isolated module during text fusion; it now incorporates contextually associated text information for fusion. Additionally, to extend the memory duration of LSTM for textual information\cite{lstm+att}, we incorporate shuffle attention by shuffling the spatial and channel features, respectively. Thus, the fusion block is enhanced to pay attention to both text and image information, resulting in a better fusion effect. The contributions of this paper include:
\begin{itemize}
    \item We propose a recurrent affine transformation module based on LSTM jump-connected feature layers, so that the fused textual information in different layers has semantic relations in the global textual information, making the fusion effect better.
    \item We introduce shuffle Attention between every two loop affine transform modules. This approach simulates the "learn-review" mode in the learning process of biological behavior. It allows the learner to suppress the forgetting of textual information and maintains the stable transfer of knowledge over time.
    \item We conducte extensive comparative experiments and visualizations to demonstrate that the proposed model improves the visual quality and evaluation metrics of the CUB, Oxford and CelebA-tiny datasets.
\end{itemize}

\begin{figure*}[!t]
\centering
\includegraphics[width=6in]{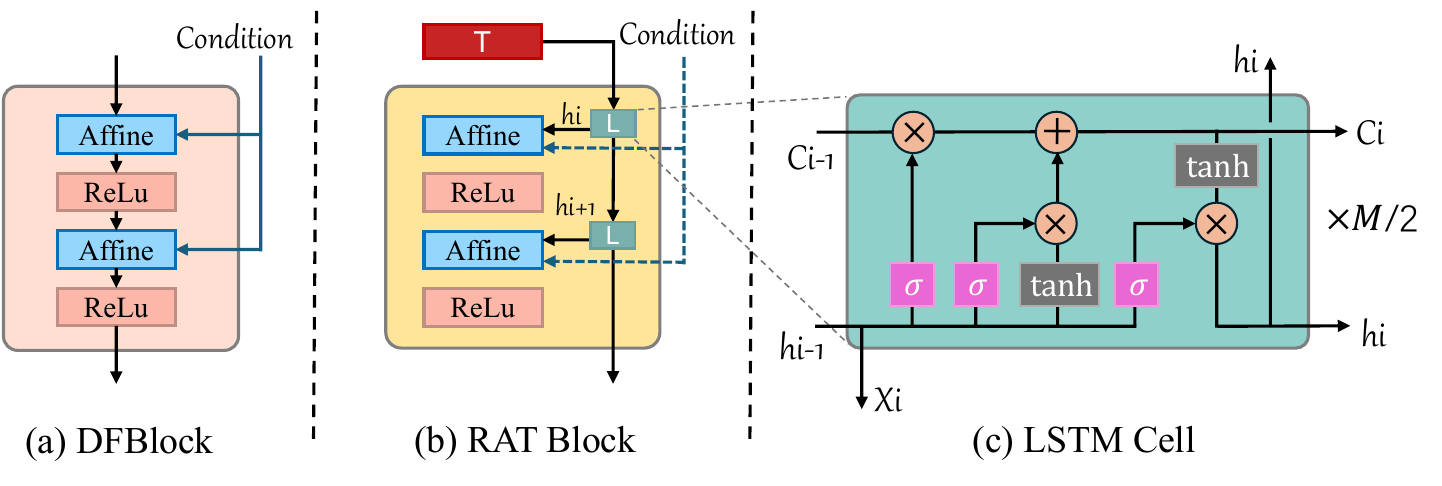}
\caption{RAT-Block structure. (a) DFBlock in Baseline (b) RAT Block, where the L-module is an LSTM cell. (c) LSTM Cell.}
\label{rat}
\end{figure*}

\section{Related Work}

\subsection{Text to Image Synthesis}

Text-to-image synthesis is a key task in conditional image synthesis. With the advancement of generative artificial intelligence models, an increasing number of generative models have been proposed, with Generative Adversarial Networks (GANs) notably achieving excellent results in this domain. Conditional GAN (CGAN)\cite{mirza2014conditional} was the first to propose a GAN with conditionals, roughly fusing textual information by connecting textual features with noise vectors. Dumoulin et al.\cite{dumoulin2016learned} proposed a more advanced fusion method, Conditional Instance Normalization (CIN), which uses adaptive mean and variance to control the image style. CIN and its variants have been widely utilized in recent research efforts. For instance, Karras et al.\cite{karras2019style} introduced StyleGAN, which leverages CIN for text-to-image synthesis. For high-resolution text-to-image synthesis, Xu et al.\cite{xu2018attngan} proposed AttnGAN, driven by attention mechanisms for more fine-grained synthesis. Tao et al.\cite{tao2022df} proposed DF-GAN, a single-stage high-resolution generative model. Zhou et al.\cite{zhou2022towards} introduced LAFITE, which utilizes the CLIP text-image contrast loss for text-to-image training, resulting in significant improvements in evaluation metrics. GALIP\cite{tao2023galip}, proposed by Tao et al., also utilizes CLIP for text-to-image synthesis. It significantly outperforms previous models in evaluation metrics including inference speed, synthesized image quality, and model parameter count.

\subsection{Attention in Text to Image}

Attention mechanisms are widely employed in various deep learning fields, and text-to-image synthesis is no exception. As mentioned earlier, AttnGAN achieved impressive results in generating high-resolution images. Lee et al.\cite{Lee_2018_ECCV} proposed a stacked cross-attention mechanism to identify comprehensive alignments and infer image-text similarity using image regions and words in sentences as context. Ye et al.\cite{ye2023recurrent} combined spatial attention in a discriminator to ensure semantic consistency between images and text. Besides its extensive use in GAN-based text-to-image synthesis, attention mechanisms are also prevalent in other generative models. Chefer et al.\cite{chefer2023attend} applied refined cross-attention units in a stable diffusion model to enhance the accuracy of text content synthesis.

Our study is inspired by the work of Ye et al. \cite{ye2023recurrent}, and based on the feature that LSTMs are prone to forgetting and losing some textual information during prolonged learning \cite{arora2019does}, we introduce shuffle attention to between two RAT Blocks, which is used to enhance textual memory for context and slow down forgetting.

\section{Method}

This paper proposes a generative adversarial CLIP model based on recurrent affine transform (RATLIP). To enhance the ability of the isolated fusion block to capture global text information and achieve better text fusion, we propose: (i) a novel recurrent affine transform module (RAT Block) to improve the fusion of text and images. We introduce shuffle Attention between RAT Blocks to mitigate LSTM forgetting. In this section, we first present the overall framework of RATLIP and then elaborate on the rationale behind RAT Block and shuffle Attention combination.

\subsection{Overall Framework of RATLIP} 
As shown in Figure \ref{all}.
Our proposed RATLIP improves on the framework of GALIP\cite{tao2023galip} as a baseline. The overall consists of a pre-trained CLIP text encoder, a generator (G), and a discriminator (D).

In the generator, a pre-trained CLIP text encoder encodes the input text description into a sentence vector T\cite{SMC2023-gan}. The noise vector Z is sampled from a Gaussian distribution, allowing for more diversity in the synthesized content. First, the vectors T and Z are input to the RAT Bridge, which contains RAT blocks (RATBLKs), and a convolutional module (SAConv) with Shuffle Attention is associated with every two RATBLKs. The RAT Bridge transforms the input vectors into bridges conforming to the frozen CLIP-ViT feature value $F$ and passes it into CLIP-BLK. Finally, the image generator (image-G) synthesizes high-quality images from the vectors T, Z and the bridge features $F$ coming out of CLIP-BLK.

The discriminator is based on the frozen CLIP-ViT, and it includes a paired discriminator (Mate-D)\cite{tao2023galip}.CLIP-ViT in CLIP-BLK converts images into image features using a convolutional layer and a series of converter blocks. Mate-D consists of two BLKs in yellow (FE Extractor) and purple (Referee) in Fig. \ref{all} FE Extractor collects bridge feature $F$ from CLIP-BLK. The adversarial loss is predicted by $F$ and sentence vector T in Referee. By distinguishing between synthetic and real images, the discriminator facilitates the generator to synthesize higher quality images.

\subsection{Recurrent Affine Transformations}
CAT is a multilayer perceptron used to independently predict batch-by-batch statistics between neighboring layers, without access to global textual information by other layers. In this section, we address this issue by employing Recurrent Affine Transformation (RAT) to improve the consistency of global textual information across fusion blocks in different layers.

The RAT first performs a channel scaling operation on the image feature vector c using the scaling parameter, and then applies a range of channel shift operations on c using the shift parameter. This process can be viewed as:
\begin{equation}
    Affine \left(c \mid h_{i}\right)=\gamma_{i} \cdot c+\beta_{i}\\
\end{equation}
where $h_i$ is the number of hidden states of the RNN,$\gamma$ and $\beta$ are the prediction parameters of the single-layer perceptron machine conditioned on $h_i$ with the following expressions:
\begin{equation}
      \gamma=\operatorname{MLP}_{1}\left(h_{i}\right), \quad \beta=\operatorname{MLP}_{2}\left(h_{i}\right)
\end{equation}
the specific implementation is shown in Fig. \ref{rat}: Fig. \ref{rat}-a depicts the text fusion module DFBlock in the baseline. The "condition" represents the conditional affine transform structure (CAT) of the multilayer perceptron, which returns two prediction parameters through learning. Fig. \ref{rat}-b illustrates the RAT Block we designed, which exhibits clear advantages over the baseline in terms of structure. By utilizing the LSTM Cell shown in Fig. \ref{rat}-c for jump connection and weight contribution, the text information between fusion blocks remains consistent without requiring the same DFBlock for text fusion based on batch statistics. This significantly enhances the degree of text fusion.
\begin{table*}[!t]
\caption{Our model compares results with current state-of-the-art methods on the test set of datasets CUB, Oxford, CelebA-tiny.}
\centering
\renewcommand\arraystretch{1.5}
\begin{tabular}{lccccc}
\hline
                                   & \multicolumn{2}{c}{\textbf{FID↓}}                                                & \multicolumn{3}{c}{\textbf{CS↑}}                          \\ \cline{2-6} 
\multirow{-2}{*}{\textbf{Method}} & \textbf{CUB}                                   & \multicolumn{1}{l}{\textbf{CelebA-tiny}} & \textbf{CUB}                          & \textbf{Oxford}                       & \multicolumn{1}{l}{\textbf{CelebA-tiny}} \\ \hline
\textbf{AttnGAN\cite{xu2018attngan}}                   & 23.98                                 & 125.98                                   & -                                     & -                                     & 21.15                                    \\
\textbf{LAFITE\cite{zhou2022towards}}                    & 14.58                                 & -                                        & 31.25                                 & -                                     & -                                        \\
\textbf{DF-GAN\cite{tao2022df}}                    & 14.81                                 & 137.6                                    & 29.20                                 & 26.67                                 & 24.41                                    \\
\textbf{GALIP\cite{tao2023galip}}            & {\color[HTML]{FF0000} 10}             & 94.45                                    & 31.60                                 & 31.77                                 & 27.95                                    \\
\textbf{RATLIP(Ours)}             & {\color[HTML]{0070C0} \textbf{13.28}} & {\color[HTML]{FF0000} \textbf{81.48}}    & {\color[HTML]{FF0000} \textbf{32.03}} & {\color[HTML]{FF0000} \textbf{31.94}} & {\color[HTML]{FF0000} \textbf{28.91}}    \\ \hline
\label{compare}
\end{tabular}
\end{table*}
We use RNNs, specifically LSTM, to model the temporal structure within a RAT block to incorporate textual information globally.Therefore, we use the widely used LSTM. The initial state of the LSTM is computed based on the noise vector Z. The initial state of the LSTM is determined by the noise vector Z:
\begin{equation}
    h_{0}=\operatorname{MLP}_{3}(z), \quad c_{0}=\operatorname{MLP}_{4}(z)
\end{equation}
The update rule for the RAT constructed by LSTM is:
\begin{equation}
    \left(\begin{array}{c}\mathbf{i}_{\mathbf{t}} \\ \mathbf{f}_{t} \\ \mathbf{o}_{t} \\ u_{t}\end{array}\right)=\left(\begin{array}{c}\sigma \\ \sigma \\ \sigma \\ \tanh \end{array}\right)\left(T\binom{s}{h_{t-1}}\right)
\end{equation}
\begin{equation}
    \mathbf{c}_{t}=\mathbf{f}_{t} \odot \mathbf{c}_{t-1}+\mathbf{i}_{t} \odot u_{t}
\end{equation}
\begin{equation}
    h_{t}=\mathbf{o}_{t} \odot \tanh \left(\mathbf{c}_{t}\right)
\end{equation}
\begin{equation}
    \gamma_{t}, \beta_{t}=\operatorname{MLP}_{1}^{\mathrm{t}}\left(h_{t}\right), \operatorname{MLP}_{2}^{\mathrm{t}}\left(h_{t}\right)
\end{equation}
where $i_t$,$f_t$,$0_t$ are the input gate, oblivion gate and output gate, respectively. $\sigma,\tanh$ are the sigmoid and tanh activation functions, respectively, and the affine transformation is:
$$L^{D+d} \rightarrow L^{d}$$
where d is the dimension of the text embedding and D is the number of RNN hidden state units.

\subsection{Suppression of Text forgetting}

According to the feature that LSTM is prone to information forgetting in long time learning \cite{arora2019does}, we introduce shuffle attention (SA) to between two RAT Blocks, which is used to enhance the memory of the text to the context and slow down the information forgetting.

Convolutional Block Attention Module (CBAM) aims to capture both pairwise pixel-level relationships and inter-channel dependencies. Spatial and channel attention aim for optimal results, but they inevitably introduce a significant computational burden to the model. Shuffle Attention (SA) can better address this issue. Shuffle Attention (SA) employs a shuffle rule where incoming parameters are grouped fairly. This interaction shuffles spatial and channel information separately, and the resulting groups are then re-fused, leading to a rich information representation.

As depicted in the internal structure of RAT-Bridge in Figure \ref{all}. As we will discuss below, since each LSTM cell stores 64 hidden layers, the deeper layers of the network may lose part of their parameters during gradient descent, a phenomenon referred to as "biological oblivion". We add SA after the first RATBLK to allow the information to pass through the learner first and then focus on it. This approach is analogous to behavioral learning in biology: first learning (RATBLK learning) and then reviewing (shuffle attention), iterating in this manner to maintain long-term stable knowledge transfer. This enables our model to mitigate the forgetting of textual information.

\subsection{Objective Function}
The RATLIP model will predict the adversarial loss based on the extracted information features and sentence vectors with the following formula:
\begin{equation}
    \begin{aligned} L_{D}= & -\mathbb{E}_{x \sim \mathbb{P}_{r}}[\min (0,-1+D(C(x), T))] \\ & -(1 / 2) \mathbb{E}_{G(z, T) \sim \mathbb{P}_{g}}[\min (0,-1-D(C(G(z, T)), T))] \\ & -(1 / 2) \mathbb{E}_{x \sim \mathbb{P}_{m i s}}[\min (0,-1-D(C(x),T))] \\ & +k \mathbb{E}_{x \sim \mathbb{P}_{r}}
    \\ &\left[\left(\left\|\nabla_{C(x)} D(C(x), T)\right\|+\left\|\nabla_{T} D(C(x), T)\right\|\right)^{p}\right], \\ L_{G}= & -\mathbb{E}_{G(z, T) \sim \mathbb{P}_{g}}[D(C(G(z, T)), T)] \\ & -\lambda \mathbb{E}_{G(z, T) \sim \mathbb{P}_{g}}[S(G(z, T), T)],\end{aligned}
\end{equation}
where $z$ is the noise vector sampled from a Gaussian distribution; $T$ is the sentence vector; $G$ is the generator function; $D$ is the discriminator function; $C$ is the CLIP-ViT frozen in the CLIP-based discriminator; $S$ denotes the cosine similarity computed in the CLIP score; $k$ and $p$ are two hyper-parameters of the gradient penalization; $\lambda$ is the value of the FID; and $P_g,P_r,P_{mis}$ denote synthetic data distributions and real data distributions and mismatched data distributions, respectively. denote the synthetic data distribution, the real data distribution and the mismatched data distribution.

\section{Experiment}
\subsection{Dataset}
For our experiments, we selected popular datasets in this field: CUB\cite{UCSD}, Oxford\cite{nilsback2008automated}, and CelebA-tiny. CelebA-tiny is based on the CelebAMask-HQ\cite{ CelebAMask-HQ}, with 10,000 photos randomly selected for text-image matching. Of these, 8,000 are in the training set and 2,000 are in the test set. Additionally, the CUB dataset comprises 200 different categories, totaling 11,788 bird images. The Oxford dataset contains 102 different categories, totaling 8,189 images of flowers. Each of these three datasets includes 10 sentences describing each image.
\subsection{Training and Evaluation Details}
We chose the ViT-B/32\cite{radford2021learning} model as the model for CLIP to use in our RATLIP. The learning rates of the generator and the discriminator are: 0.0001 and 0.0004 respectively, and Adam is chosen as the optimizer for both networks. Continuing the recent text-to-image work\cite{tao2023galip}\cite{tao2022df}\cite{zhou2022towards}\cite{ye2023recurrent}, the model evaluation metrics we use are FID\cite{2017GANs} and CLIP- Score(CS)\cite{wallacesupplementary}. All experiments were done in $3\times 3090$ GPUs.

\subsection{Quantitative Evaluation}
\textbf{Comparison Experiment.} To evaluate the performance of RATLIP, we compare it with several current state-of-the-art methods, all of which have achieved impressive results\cite{tao2023galip}\cite{tao2022df}\cite{xu2018attngan}\cite{zhou2022towards}. The results are shown in Table \ref{compare}. Under the FID metric, CeleA-tiny ranked first, while CUB ranked second (a lower d-score indicates a higher ranking). CeleA-tiny achieved state-of-the-art (SOTA) performance. Under the CS metric, all three datasets improved by approximately 0.78 to 0.96 compared to the second place score.

\begin{figure}[!thb]
\centering
\includegraphics[width=3.3in]{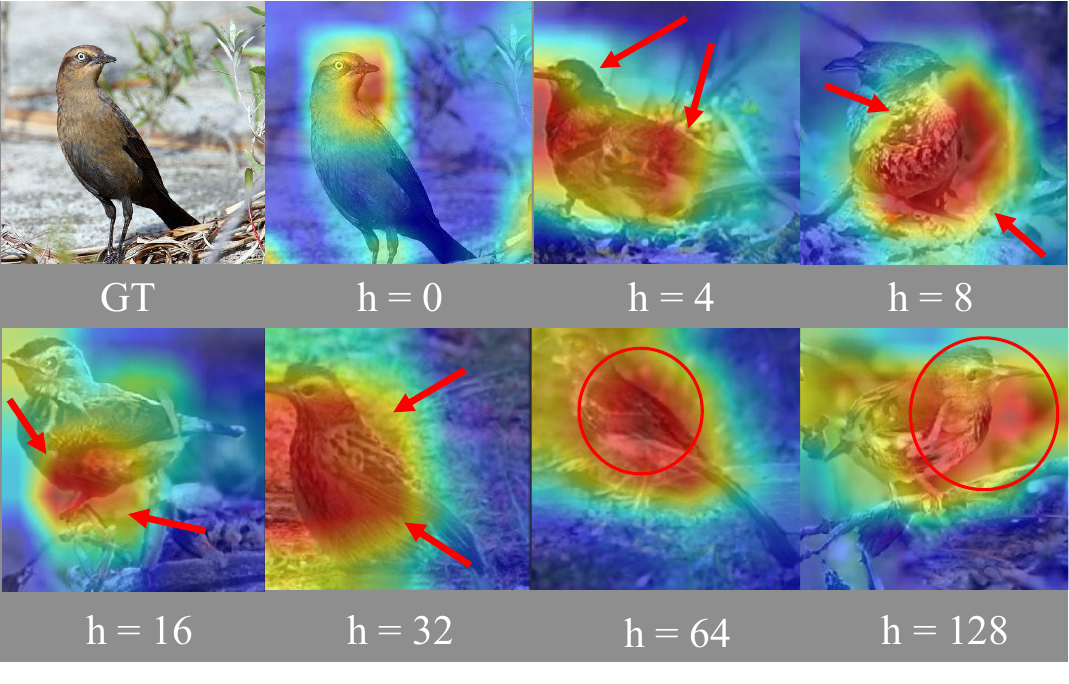}
\caption{Image synthesized by LSTM with different hidden layers h in CUB dataset visualization result under CAM. The image starts from groud truth(h=0) and ends at h=128.}
\label{cam-lstm}
\end{figure}

\textbf{Parametric Analysis.}
LSTM Cell has hidden layer $h_i$,Values that are too high or too low will have a negative impact on the model. Therefore, we conducted parameter analysis experiments to explore the effect of $h=0, 4, 8, 16, 32, 64, 128$ on the total number of model parameters and CS values, respectively. In Table \ref{lstm}, the results of our model are highlighted in red. As the number of network layers increases, the CS score shows an upward trend, but at the same time, the number of computed parameters increases rapidly. To select a suitable $h$ value, we used Grad-CAM\cite{Gradcam} to visualize the synthesized photo, as shown in Fig. \ref{cam-lstm}. Notice that the significant feature regions are shown in red, from Fig. $h=0$ to $h =32$, the red regions are distributed to all areas of the target, either not concentrated or off the torso, and cannot cover the target well, and at $h=64$, the The red region completely covers the target, and the same is true for $h =128$, but the area covered is larger. In summary, it can be tentatively concluded that the most ideal hidden layer is $h=64$. 

\begin{figure}[!htb]
\centering
\includegraphics[width=3.3in]{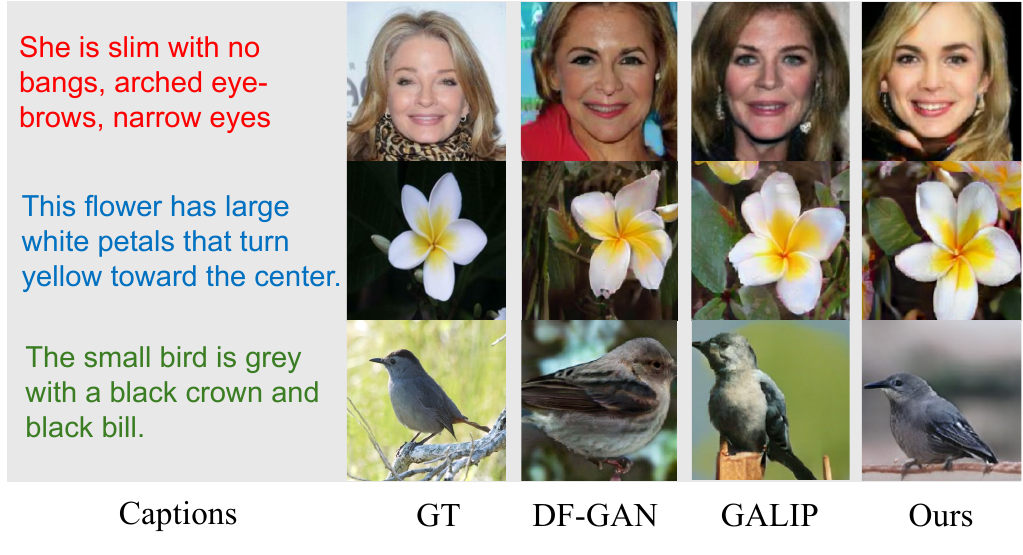}
\caption{Visualization comparison of the dataset CelebA-tiny,Oxford,CUB and current state-of-the-art models.}
\label{vision}
\end{figure}

\subsection{Qualitative Evaluation}
\textbf{Visualization Results.}
Fig.\ref{vision} presents a visual comparison between the current state-of-the-art model and our model. The images in the last column represent the results of our model. It is evident that our model has a significant advantage in terms of color fidelity, as the human hair color, flower petal color, and bird feather color all closely match the original colors.
\begin{figure}[!htb]
\centering
\includegraphics[width=3.4in]{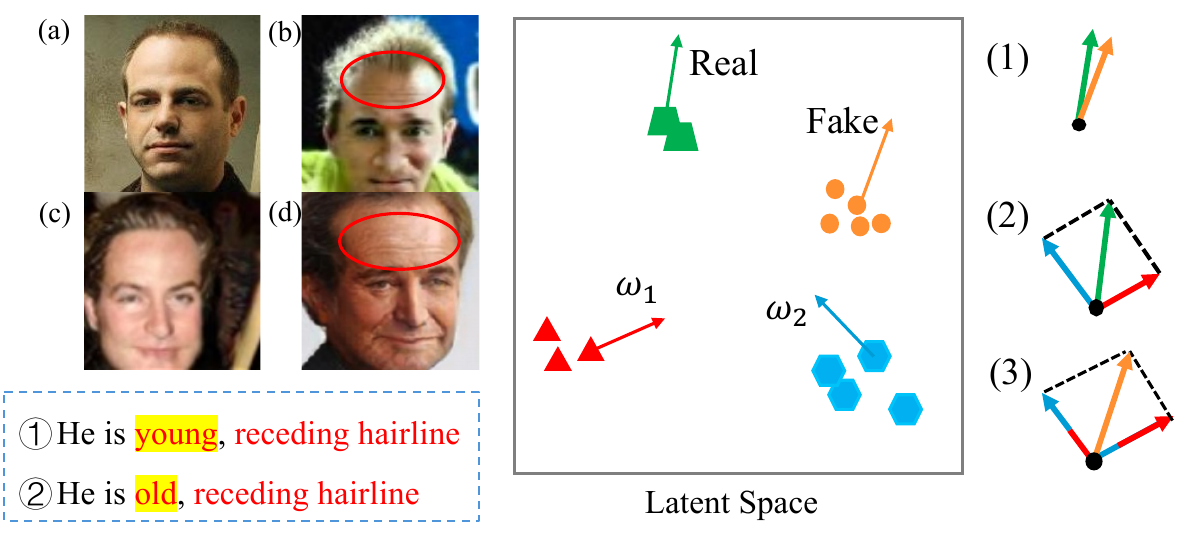}

\caption{Semantic spatial feature qualitative analysis. The face images in (a),(b),(c),(d) are GT, baseline, Ours(young),Ours(Old). On the right, there is a latent space containing vectors representing four image features. The rule states that when you add vectors together, the result is a vector in another feature.}
\label{space}
\end{figure}

\textbf {Latent Space Visualization Analysis.}
During our visualization exploration, we observed some interesting phenomena. In Figure \ref{space}, we used GALIP (baseline) and our RATLIP to synthesize the first caption: "He is young, receding hairline" to obtain (b) and (c). In the baseline, "young" is overridden by "receding hairline", resulting in wrinkles on the forehead of the face in the figure. In contrast, our model generates a more semantically appropriate image without wrinkles, and the hairline is not affected. We hypothesize that the image features in the baseline cannot be completely separated, and some of the textual information causes the feature vectors to fuse in a confusing manner, leading to deviations from the real direction in the final result. To test this hypothesis, we synthesized a second caption on RATLIP, replacing "young" with "old", and our results immediately showed wrinkles, as seen in Figure \ref{space}-(d).

This phenomenon can be explained in the latent space shown in Figure \ref{space}. Each color block corresponds to a type of feature: green represents real image features (vector real), and orange represents fake image features generated by the model (vector fake). In Fig. \ref{space}-(1), the smaller the angle between real and fake vectors, the closer the generated image is to the real image. In Fig. \ref{space}-(2) and Fig. \ref{space}-(3) of our model and the baseline, respectively, it can be seen that in Fig. \ref{space}-(3), there is a relatively chaotic feature fusion, leading to a significant deviation in the generated image, resulting in inconsistency with the above-mentioned graphic.

\begin{table}[]
\caption{LSTM Hidden Layers Validation for CUB Dataset}
\centering
\renewcommand\arraystretch{1.5}
\begin{tabular}{lcccc}

\hline
                                       &                                       & \multicolumn{2}{c}{\textbf{Params}}                                             &                                      \\ \cline{3-4}
\multirow{-2}{*}{\textbf{RATLIP(CUB)}} & \multirow{-2}{*}{\textbf{CS}} & \textbf{GPs(B)}                        & \textbf{TPs(B)}                        & \multirow{-2}{*}{\textbf{ATT}} \\ \hline
\textbf{RAT(hide=0)}                   & 31.62                                 & 131.01                                 & 160.39                                 &                                      \\
\textbf{hide=4}                        & 30.81                                 & 132.09                                 & 161.47                                 & \checkmark                                    \\
\textbf{hide=8}                        & 31.38                                 & 132.27                                 & 161.65                                 & \checkmark                                    \\
\textbf{hide=16}                       & 31.02                                 & 132.57                                 & 161.95                                 & \checkmark                                    \\
\textbf{hide=32}                       & 31.89                                 & 133.25                                 & 162.63                                 & \checkmark                                    \\
\textbf{hide=64  }                              & {\color[HTML]{FF0000} \textbf{32.03}} & {\color[HTML]{FF0000} \textbf{134.68}} & {\color[HTML]{FF0000} \textbf{164.06}} & \checkmark                                    \\
\textbf{hide=128}                      & 32.12                                 & 137.93                                 & 167.31                                 & \checkmark                                    \\ \hline
\label{lstm}
\end{tabular}
\end{table}

\subsection{Ablation Experiment}
In order to verify the validity of each component of our model, we conduct ablation experiments on CUB, Oxford, and CelebA-tiny datasets respectively, and use the CS score as the evaluation index. It can be seen in Table \ref{ablation}:
\begin{itemize}
    \item \textbf{Effect of RAT Block in generator.}Our model improves over baseline by 0.02, 0.06 on the datasets CUB, Oxford, and CelebA-tiny, respectively, but the metrics drop to 27.63 on the CelebA-tiny dataset.Initially, we hypothesize that the effect of the deeper network enhancement is is not large or the metrics drop has a lot to do with the fact that the LSTM learner appears to be oblivious.
    \item \textbf{Effect of shuffle attention (SA) in every two RAT Blocks.} This improvement leads to a substantial improvement in the CS metrics, reaching the best results respectively as shown in the last row of Table \ref{ablation}. It is demonstrated that SA attention allows the bridge feature extractor to focus more accurately on the input text and image information. Suppression of LSTM learner forgetfulness is achieved.
\end{itemize}

\begin{table}[]
\caption{Ablation Experiment}
\centering
\renewcommand\arraystretch{1.5}
\begin{tabular}{lcccc}
\hline
  & \multicolumn{3}{c}{\textbf{CS↑}}    &                                \\ \cline{2-4}
\multirow{-2}{*}{\textbf{Method}}               & \textbf{CUB}                 & \textbf{Oxford}              & \multicolumn{1}{l}{\textbf{CelebA-tiny}} & \multirow{-2}{*}{\textbf{ATT}} \\ \hline
\textbf{Baseline\cite{tao2023galip}}                       & 31.60                        & 31.77                        & 27.95                                    & ×                              \\
\textbf{RAT}                            & 31.62                        & 31.83                        & 27.63                                    & ×                              \\
\textbf{RAT+ATT} & {\color[HTML]{FF0000} \textbf{32.03}} & {\color[HTML]{FF0000} \textbf{31.94}} & {\color[HTML]{FF0000} \textbf{28.91}}             & \checkmark                              \\ \hline
\label{ablation}
\end{tabular}
\end{table}
 
\section{Result}
In this paper, we propose an adversarial CLIP-based approach for text-to-image synthesis called Recurrent Affine Transformation for Text-to-Image Synthesis (RATLIP). Compared to previous text fusion approaches, RATLIP maintains the consistency of text information across different layers of fusion blocks, enables interaction between isolated fusion blocks, and achieves superior text fusion. Additionally, pre-trained CLIP addresses issues such as the lack of diverse synthetic images in generative adversarial networks. Including SA attention between every two RATs simulates the response of the biological nervous system to inhibit forgetting. This ensures that information is not lost significantly in the deeper layers of the network. Finally, extensive experiments with RATLIP demonstrate that the proposed framework enhances the quality of image generation, and we hope that this work will provide valuable guidance for future research.
\bibliography{my} 

\begin{thebibliography}{10}

\bibitem{Xie_2023_ICCV}
Jinheng Xie, Yuexiang Li, Yawen Huang, Haozhe Liu, Wentian Zhang, Yefeng Zheng, and Mike~Zheng Shou.
\newblock Boxdiff: Text-to-image synthesis with training-free box-constrained diffusion.
\newblock In {\em Proceedings of the IEEE/CVF International Conference on Computer Vision (ICCV)}, pages 7452--7461, 2023.

\bibitem{2024diffusion}
Gan Sun, Wenqi Liang, Jiahua Dong, Jun Li, Zhengming Ding, and Yang Cong.
\newblock Create your world: Lifelong text-to-image diffusion.
\newblock {\em IEEE Transactions on Pattern Analysis and Machine Intelligence}, 2024.

\bibitem{xue2024}
Zeyue Xue, Guanglu Song, Qiushan Guo, Boxiao Liu, Zhuofan Zong, Yu~Liu, and Ping Luo.
\newblock Raphael: Text-to-image generation via large mixture of diffusion paths.
\newblock {\em Advances in Neural Information Processing Systems}, 36, 2024.

\bibitem{tao2023galip}
Ming Tao, Bing-Kun Bao, Hao Tang, and Changsheng Xu.
\newblock Galip: Generative adversarial clips for text-to-image synthesis.
\newblock In {\em Proceedings of the IEEE/CVF Conference on Computer Vision and Pattern Recognition}, pages 14214--14223, 2023.

\bibitem{dumoulin2016learned}
Vincent Dumoulin, Jonathon Shlens, and Manjunath Kudlur.
\newblock A learned representation for artistic style.
\newblock {\em arXiv preprint arXiv:1610.07629}, 2016.

\bibitem{karras2019style}
Tero Karras, Samuli Laine, and Timo Aila.
\newblock A style-based generator architecture for generative adversarial networks.
\newblock In {\em Proceedings of the IEEE/CVF conference on computer vision and pattern recognition}, pages 4401--4410, 2019.

\bibitem{patashnik2021styleclip}
Or~Patashnik, Zongze Wu, Eli Shechtman, Daniel Cohen-Or, and Dani Lischinski.
\newblock Styleclip: Text-driven manipulation of stylegan imagery.
\newblock In {\em Proceedings of the IEEE/CVF international conference on computer vision}, pages 2085--2094, 2021.

\bibitem{brock2018large}
Andrew Brock, Jeff Donahue, and Karen Simonyan.
\newblock Large scale gan training for high fidelity natural image synthesis.
\newblock {\em arXiv preprint arXiv:1809.11096}, 2018.

\bibitem{tao2022df}
Ming Tao, Hao Tang, Fei Wu, Xiao-Yuan Jing, Bing-Kun Bao, and Changsheng Xu.
\newblock Df-gan: A simple and effective baseline for text-to-image synthesis.
\newblock In {\em Proceedings of the IEEE/CVF Conference on Computer Vision and Pattern Recognition}, pages 16515--16525, 2022.

\bibitem{huang2017arbitrary}
Xun Huang and Serge Belongie.
\newblock Arbitrary style transfer in real-time with adaptive instance normalization.
\newblock In {\em Proceedings of the IEEE international conference on computer vision}, pages 1501--1510, 2017.

\bibitem{lstm+att}
Bahareh Nikpour and Narges Armanfard.
\newblock Spatial hard attention modeling via deep reinforcement learning for skeleton-based human activity recognition.
\newblock {\em IEEE Transactions on Systems, Man, and Cybernetics: Systems}, 53(7):4291--4301, 2023.

\bibitem{mirza2014conditional}
Mehdi Mirza and Simon Osindero.
\newblock Conditional generative adversarial nets.
\newblock {\em arXiv preprint arXiv:1411.1784}, 2014.

\bibitem{xu2018attngan}
Tao Xu, Pengchuan Zhang, Qiuyuan Huang, Han Zhang, Zhe Gan, Xiaolei Huang, and Xiaodong He.
\newblock Attngan: Fine-grained text to image generation with attentional generative adversarial networks.
\newblock In {\em Proceedings of the IEEE conference on computer vision and pattern recognition}, pages 1316--1324, 2018.

\bibitem{zhou2022towards}
Yufan Zhou, Ruiyi Zhang, Changyou Chen, Chunyuan Li, Chris Tensmeyer, Tong Yu, Jiuxiang Gu, Jinhui Xu, and Tong Sun.
\newblock Towards language-free training for text-to-image generation.
\newblock In {\em Proceedings of the IEEE/CVF Conference on Computer Vision and Pattern Recognition}, pages 17907--17917, 2022.

\bibitem{Lee_2018_ECCV}
Kuang-Huei Lee, Xi~Chen, Gang Hua, Houdong Hu, and Xiaodong He.
\newblock Stacked cross attention for image-text matching.
\newblock In {\em Proceedings of the European Conference on Computer Vision (ECCV)}, September 2018.

\bibitem{ye2023recurrent}
Senmao Ye, Huan Wang, Mingkui Tan, and Fei Liu.
\newblock Recurrent affine transformation for text-to-image synthesis.
\newblock {\em IEEE Transactions on Multimedia}, 2023.

\bibitem{chefer2023attend}
Hila Chefer, Yuval Alaluf, Yael Vinker, Lior Wolf, and Daniel Cohen-Or.
\newblock Attend-and-excite: Attention-based semantic guidance for text-to-image diffusion models.
\newblock {\em ACM Transactions on Graphics (TOG)}, 42(4):1--10, 2023.

\bibitem{arora2019does}
Gaurav Arora, Afshin Rahimi, and Timothy Baldwin.
\newblock Does an lstm forget more than a cnn? an empirical study of catastrophic forgetting in nlp.
\newblock In {\em Proceedings of the The 17th Annual Workshop of the Australasian Language Technology Association}, pages 77--86, 2019.

\bibitem{SMC2023-gan}
Chao Guo, Yong Dou, Tianxiang Bai, Xingyuan Dai, Chunfa Wang, and Yi~Wen.
\newblock Artverse: A paradigm for parallel human–machine collaborative painting creation in metaverses.
\newblock {\em IEEE Transactions on Systems, Man, and Cybernetics: Systems}, 53(4):2200--2208, 2023.

\bibitem{UCSD}
Peter Welinder, Steve Branson, Takeshi Mita, Catherine Wah, Florian Schroff, Serge Belongie, and Pietro Perona.
\newblock Caltech-ucsd birds 200.
\newblock 2010.

\bibitem{nilsback2008automated}
Maria-Elena Nilsback and Andrew Zisserman.
\newblock Automated flower classification over a large number of classes.
\newblock In {\em 2008 Sixth Indian conference on computer vision, graphics \& image processing}, pages 722--729. IEEE, 2008.

\bibitem{CelebAMask-HQ}
Cheng-Han Lee, Ziwei Liu, Lingyun Wu, and Ping Luo.
\newblock Maskgan: Towards diverse and interactive facial image manipulation.
\newblock In {\em IEEE Conference on Computer Vision and Pattern Recognition (CVPR)}, 2020.

\bibitem{radford2021learning}
Alec Radford, Jong~Wook Kim, Chris Hallacy, Aditya Ramesh, Gabriel Goh, Sandhini Agarwal, Girish Sastry, Amanda Askell, Pamela Mishkin, Jack Clark, et~al.
\newblock Learning transferable visual models from natural language supervision.
\newblock In {\em International conference on machine learning}, pages 8748--8763. PMLR, 2021.

\bibitem{2017GANs}
Martin Heusel, Hubert Ramsauer, Thomas Unterthiner, Bernhard Nessler, and Sepp Hochreiter.
\newblock Gans trained by a two time-scale update rule converge to a local nash equilibrium.
\newblock 2017.

\bibitem{wallacesupplementary}
Bram Wallace, Akash Gokul, and Nikhil Naik.
\newblock Supplementary material for edict: Exact diffusion inversion via coupled transformations.

\bibitem{Gradcam}
Ramprasaath~R. Selvaraju, Michael Cogswell, Abhishek Das, Ramakrishna Vedantam, Devi Parikh, and Dhruv Batra.
\newblock Grad-cam: Visual explanations from deep networks via gradient-based localization.
\newblock In {\em 2017 IEEE International Conference on Computer Vision (ICCV)}, pages 618--626, 2017.

\end{thebibliography}
\balance

\end{document}